\documentclass[runningheads]{llncs}
\usepackage[T1]{fontenc}
\usepackage{graphicx,verbatim}
\usepackage{booktabs}
\usepackage{multirow}
\usepackage[table]{xcolor}
\usepackage{colortbl}
\usepackage{amsmath,amssymb,mathtools}
\usepackage{enumitem}
\usepackage{xcolor}
\usepackage{hyperref}
\usepackage{cleveref}
\usepackage{algorithm}
\usepackage{algpseudocode}

\newcommand{\corr}{\textsuperscript{\ensuremath{\dagger}}}

\DeclareMathOperator{\softmax}{softmax}
\DeclareMathOperator{\CE}{CE}

\newcommand{\Loss}{\mathcal{L}}

\newsavebox\CBox
\newcommand*\textBF[1]{\sbox\CBox{#1}\resizebox{\wd\CBox}{\ht\CBox}{\textbf{#1}}}

\begin{document}
\title{When, Where, and How: Adaptive Binning for Tabular Self-Supervised Learning}
\titlerunning{Adaptive Binning for Tabular SSL}

\author{
Daehwan Kim\inst{1}\orcidID{0009-0006-7856-7850} \and
Haejun Chung\inst{1}\corr\orcidID{0000-0001-8959-237X} \and
Ikbeom Jang\inst{2}\corr\orcidID{0000-0002-6901-983X}
}

\authorrunning{D. Kim et al.}

\institute{
Hanyang University, Seoul, South Korea\\
\email{\{officialhwan, haejun\}@hanyang.ac.kr}
\and
Hankuk University of Foreign Studies, Yongin, South Korea\\
\email{ijang@hufs.ac.kr}
}

\maketitle

\begingroup
\renewcommand{\thefootnote}{\ensuremath{\dagger}}
\footnotetext[0]{Corresponding authors.}
\endgroup

\begin{abstract}
Medical tabular data are ubiquitous in clinical research, but deep learning for tables remains underexplored because reliable labels often require costly expert adjudication, even though structured clinical variables are routinely available in tabular form.  Self-supervised learning can leverage these unlabeled tables, and recent binning-based pretexts offer a promising inductive bias, but existing objectives fix a single global quantile discretization and apply feature-agnostic supervision. We propose Adaptive Binning, a training-adaptive discretization pretext for tabular SSL that couples discretization to learning through a feature-wise coarse-to-fine curriculum. Motivated by the spectral bias of neural networks and the principles of curriculum learning, our method progressively refines discretization per feature upon plateau detection and selects representation-aware splits to jointly improve value-space concentration and representation-space coherence. A heterogeneity-aware objective unifies categorical reconstruction with ordinal supervision for numerical features, and experiments on public medical tabular datasets under unified evaluation protocols show consistent gains for linear probing and fine-tuning without dataset-specific discretization tuning.
We further introduce a medical tabular SSL benchmark with standardized protocols to support reproducible progress in this underexplored domain. Our code is available at \href{https://github.com/labhai/Adaptive-Binning}{https://github.com/labhai/Adaptive-Binning}.


\end{abstract}

\section{Introduction}
Clinical trials, registries, and epidemiological studies routinely tabulate baseline characteristics, laboratory panels, graded findings, and outcomes; in a pilot review of comparative clinical trials, 99\% of articles reported baseline or outcome measures in at least one table, and 85\% reported both~\cite{holub2021toward}. This reliance on tables reflects a broader pattern where tabular data dominates structured decision systems but remains underexplored in deep learning~\cite{borisov2022deep}. Primary challenges are that tables mix categorical and numerical variables, exhibit non-smooth interactions, and lack spatial or sequential structure~\cite{gorishniy2021revisiting,grinsztajn2022treebasedmodelsoutperformdeep}. These properties favor tree ensembles such as XGBoost~\cite{chen2016xgboost} and CatBoost~\cite{prokhorenkova2018catboost}, whose recursive partitioning yields piecewise-constant functions for mixed types~\cite{shwartz2022tabular,mcelfresh2023neural}. While tabular neural architectures narrow this gap~\cite{gorishniy2021revisiting,arik2021tabnet,yan2023t2g}, deep models are most compelling when they exploit self-supervised representation learning from unlabeled data, a setting that aligns with healthcare, where labels require expert adjudication~\cite{esteva2019guide,yoon2020vime}. Nonetheless, medical self-supervision has focused on imaging and language, leaving clinical tabular data underserved.

Recent tabular SSL progress recasts binning as a pretext task~\cite{lee2024binning}, discretizing continuous features into quantile bins and reconstructing bin indices to inject a tree-like continuous-to-discrete inductive bias and harmonize supervision across heterogeneous features. However, discretization is fixed globally: a single bin count $T$ with static quantile boundaries persists throughout training, and numerical targets are fit by pointwise squared-error regression on integer indices. This feature-agnostic design neither adapts resolution as features saturate nor uses representations to localize refinement, and it provides limited support for type-aware supervision that jointly models ordinal numerical targets and categorical reconstruction. These limitations call for an SSL pretext in which discretization is training-coupled and evolves during pretraining.

Motivated by these gaps, we replace globally fixed discretization with a training-adaptive coarse-to-fine curriculum. This design mirrors the clinical progression from broad stratification to finer severity grading encoded in diagnostic criteria~\cite{mcgorry2006clinical,amin2017eighth} and aligns with neural network learning dynamics. We leverage the synergy between curriculum learning~\cite{bengio2009curriculum} and spectral bias~\cite{rahaman2019spectral}--the tendency of networks to fit coarse structures before fine details--to implement an adaptive binning strategy that gradually increases task complexity. To realize this curriculum, we develop an autoencoding-based tabular SSL framework that refines discretization feature-wise during pretraining, coupled with type-aware reconstruction. It specifies \emph{when/where/how} to refine discretization via feature-wise saturation triggers, representation-aware split selection, and a type-aware reconstruction objective for mixed categorical and ordinal numerical targets, so discretization targets evolve online during pretraining. Our contributions are:

\begin{enumerate}
    \item We propose Adaptive Binning (Fig.~\ref{figure:main}), a training-adaptive discretization pretext task for tabular SSL that replaces fixed global binning with feature-wise, coarse-to-fine refinement and explicitly specifies \emph{when}, \emph{where}, and \emph{how} discretization evolves during pretraining.
    \item We evaluate this pretext across medical tabular datasets spanning binary, nominal, and ordinal multiclass classification, and regression, using linear probing (Tab.~\ref{tab:linear_main}) and fine-tuning with multiple tabular encoders (Tab.~\ref{tab:finetune}), with a single default configuration that avoids dataset-specific tuning (Fig.~\ref{figure:loss_sensitivity}).
    \item We establish a medical tabular SSL benchmark with unified evaluation protocols (Tab.~\ref{tab:dataset_summary}), providing a reproducible foundation for progress in self-supervised learning for clinical tabular data.
\end{enumerate}

\section{Method}
\label{sec:method}
We first formalize the masking--reconstruction framework for tabular SSL and the fixed quantile-binning objective~\cite{lee2024binning} as our baseline.
We then describe Adaptive Binning, which refines discretization during pretraining and couples it with type-aware ordinal supervision for mixed categorical--numerical schemas.

\subsection{Preliminaries: Masking and Fixed Binning}
\label{sec:method:prelim}
We follow the autoencoding-based tabular SSL setup of~\cite{lee2024binning} with inputs $\mathbf{x}=[\mathbf{x}^{\mathrm{cat}},\mathbf{x}^{\mathrm{num}}]$.
During pretraining, we optionally apply feature-wise masking with probability $p_m$ and impute masked entries with a fixed constant (\emph{Const})~\cite{yoon2020vime} or an in-batch value (\emph{Random})~\cite{lee2024binning}; setting $p_m=0$ corresponds to \emph{NoMask}.
An encoder--decoder maps the corrupted input to a representation and reconstructs pretext targets, subsuming denoising/value reconstruction~\cite{vincent2008extracting} and mask detection~\cite{yoon2020vime}.
For numerical feature $n$, the fixed-binning baseline maps $x_n^{\mathrm{num}}$ to a quantile index $y^{(n)}\in\{0,\ldots,T-1\}$ using a single global bin count $T$ with static boundaries, and learns \textBF{BinRecon} by squared-error regression on $y^{(n)}$~\cite{lee2024binning}.
We consider three baseline objectives: \textBF{ValueRecon} (raw-value reconstruction), \textBF{MaskXent} (mask prediction), and \textBF{BinRecon} (fixed-$T$ quantile-index prediction).

\begin{figure}[t!]
\begin{center}
\includegraphics[width=\columnwidth]{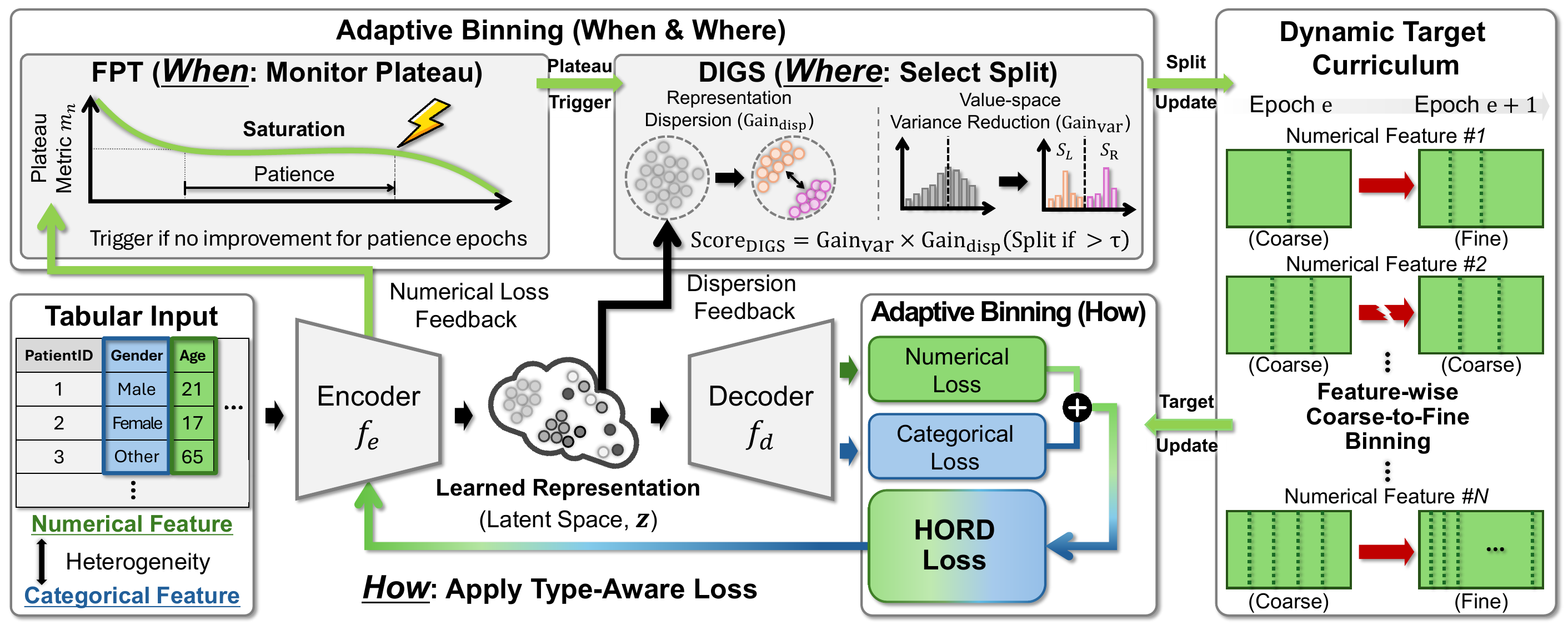}
\caption{
Overview of our proposed Adaptive Binning framework: HORD provides type-aware mixed-type reconstruction, while FPT and DIGS refine numerical binning via plateau-triggered, representation-aware splitting to form a feature-wise coarse-to-fine target curriculum; see Section~\ref{sec:method:proposed} for details.
}
\label{figure:main}
\end{center}
\end{figure}
\subsection{Proposed Method}
\label{sec:method:proposed}

For each numerical feature $n$, we maintain an adaptive discretizer $B^{(n)}$ with a feature-specific bin count $T_n$, initialized at $T_{\mathrm{init}}$ and capped at $T_{\max}$.
The bin-index target is $y^{(n)}=B^{(n)}(x^{\mathrm{num}}_n)\in\{0,\ldots,T_n\!-\!1\}$, and we refer to the per-feature schedule $\{T_n\}$ as \emph{Feature-Wise Adaptation} (FWA).

\noindent\textbf{\emph{When: Feature-Wise Plateau Trigger} (FPT).}
Numerical features vary in complexity and convergence speed, making a globally synchronized refinement schedule inefficient.
We therefore propose \emph{Feature-Wise Plateau Trigger} (FPT) to monitor each feature independently.
At the end of each epoch, we compute a feature-specific plateau metric $m_n$ from the numerical reconstruction loss, defined as the normalized weighted sum of the components in Eq.~\eqref{eq:hord_num} over the epoch; we then track the running best value $\mathrm{best}_n$ and update a patience counter $\mathrm{cnt}_n$.
If $m_n < \mathrm{best}_n - \delta$, we set $\mathrm{best}_n \leftarrow m_n$ and reset $\mathrm{cnt}_n$; otherwise, we increment $\mathrm{cnt}_n$.
Feature $n$ is eligible for splitting when $\mathrm{cnt}_n \ge \mathrm{patience}$ (set to $5$; see Fig.~\ref{figure:loss_sensitivity}) and $T_n < T_{\max}$, triggering refinement upon saturation.

\noindent\textbf{\emph{Where: Dispersion-Informed Gain-based Splitting} (DIGS).}
When FPT marks feature $n$ as ready to refine, we propose DIGS to choose \emph{which} bin to split and \emph{where} to place a new boundary by coupling variance reduction in the feature-value space with dispersion reduction in the encoder-induced representation space.
For each bin $B^{(n)}_t$ of the flagged feature, we use the within-bin median as the candidate split point, yielding a near-balanced partition that preserves equal-frequency standardization and avoids sparsity-induced target instability.
Let $S$ denote the samples in $B^{(n)}_t$, split by the median into $S_L$ and $S_R$ with weights $w_L=|S_L|/|S|$ and $w_R=|S_R|/|S|$.
For any statistic $g(\cdot)$, define the split-induced reduction as $\Delta_g(S\!\to\!S_L,S_R)=g(S)-w_L g(S_L)-w_R g(S_R)$, and set the value-space gain to $\mathrm{Gain}_{\mathrm{var}}=\Delta_{\mathrm{Var}}(S\!\to\!S_L,S_R)$~\cite{breiman2017classification}.
Since variance reduction alone is agnostic to the encoder-induced representation space, we additionally quantify within-subset coherence~\cite{de2016entropy}, which we propose to compute from embeddings of uncorrupted inputs, $\mathbf{z}_i=f_\theta(\mathbf{x}_i)$ with $\hat{\mathbf{z}}_i=\mathbf{z}_i/\|\mathbf{z}_i\|$, and define $\mathrm{Disp}(S)=\bigl|\log\!\bigl(\epsilon+\|\frac{1}{|S|}\sum_{i\in S}\hat{\mathbf{z}}_i\|^2\bigr)\bigr|$ with $\epsilon>0$, yielding $\mathrm{Gain}_{\mathrm{disp}}=\Delta_{\mathrm{Disp}}(S\!\to\!S_L,S_R)$.

We define the split score as
\begin{equation}
  \mathrm{Score}_{\mathrm{DIGS}}(S\!\to\!S_L,S_R)=\mathrm{Gain}_{\mathrm{var}}\cdot \mathrm{Gain}_{\mathrm{disp}}.
  \label{eq:digs}
\end{equation}
We split only if $\mathrm{Gain}_{\mathrm{var}}>0$, $\mathrm{Gain}_{\mathrm{disp}}>0$, and $\mathrm{Score}_{\mathrm{DIGS}}>\tau$ ($\tau=10^{-4}$; see Fig.~\ref{figure:loss_sensitivity}).
At each refinement event for feature $n$, we score all bins and apply all qualifying splits in parallel, potentially inserting multiple boundaries per trigger.
After refinement, we reset the FPT statistics for feature $n$.

\noindent\textbf{\emph{How: Heterogeneity-aware ORDinal Loss} (HORD).}
We propose HORD, a type-aware reconstruction objective that unifies nominal supervision for categorical features with ordinal, distribution-aware supervision for numerical features.
Mapping a continuous value to an ordered bin index yields an ordinal surrogate that preserves ordering and local proximity; thus, numerical reconstruction should penalize errors by bin distance rather than treat bins as nominal classes~\cite{lee2024binning}.
We supervise categorical features with cross entropy, $\Loss_{\mathrm{cat}}^{(c)}=\CE(\boldsymbol{\ell}^{(c)},y^{(c)})$.
For numerical feature $n$, we reconstruct a distribution over $T_n$ ordered bins with logits $\boldsymbol{\ell}^{(n)}\in\mathbb{R}^{T_n}$ and probabilities $\mathbf{p}^{(n)}=\softmax(\boldsymbol{\ell}^{(n)})$ for target index $y^{(n)}$.
We use soft ordinal targets~\cite{diaz2019soft} $q_t=\frac{\exp\!\bigl(-(t-y^{(n)})^2\bigr)}{\sum_{k=0}^{T_n-1}\exp\!\bigl(-(k-y^{(n)})^2\bigr)}$ and supervise numerical reconstruction with soft-target cross entropy (SORD).
We augment SORD with mean--variance regularization~\cite{pan2018mean} on $\mathbf{p}^{(n)}$ using $\mu^{(n)}=\sum_t p_t^{(n)}\,t$ and $\sigma^{2(n)}=\max\!\bigl(0,\sum_t p_t^{(n)}\,t^2-(\mu^{(n)})^2\bigr)$:
\begin{equation}
  \Loss_{\mathrm{num}}^{(n)}
  =
  w_{\mathrm{SORD}}
  \Bigl(-\sum_{t} q_t\log p^{(n)}_t\Bigr)
  \;+\;
  w_{\mathrm{mse}}\bigl(\mu^{(n)}-y^{(n)}\bigr)^2
  \;+\;
  w_{\mathrm{var}}\,\sigma^{2(n)}.
  \label{eq:hord_num}
\end{equation}
We fix $w_{\mathrm{SORD}}=10$, $w_{\mathrm{mse}}=0.1$, and $w_{\mathrm{var}}=0.001$ in all experiments (see Fig.~\ref{figure:loss_sensitivity} for sensitivity).
Finally, we average losses within each feature type and weight by feature counts, yielding uniform feature-wise weighting under varying compositions:
\begin{equation}
  \Loss_{\mathrm{HORD}}
  =
  \frac{C}{C+N}\;\frac{1}{C}\sum_{c=1}^{C}\Loss_{\mathrm{cat}}^{(c)}
  \;+\;
  \frac{N}{C+N}\;\frac{1}{N}\sum_{n=1}^{N}\Loss_{\mathrm{num}}^{(n)}.
  \label{eq:hord}
\end{equation}

\begin{table*}[t!]
\centering
\caption{Dataset summary with full names, abbreviations, task type, number of classes, presence of missing values (Missing), instances (\#Inst.), features (\#Feat.), and dataset-specific evaluation configuration (batch size and MLP width/depth).}
{\fontsize{8pt}{9.6pt}\selectfont
\begin{tabular}{l l c c c  c c c}
\toprule
\multicolumn{1}{c}{Dataset} & 
\multicolumn{1}{c}{Task} & 
\multirow{2}{*}{\#Inst.} & 
\multirow{2}{*}{\#Feat.} & 
\multirow{2}{*}{Missing} & 
Batch & 
\multicolumn{2}{c}{MLP} \\
\cmidrule(l{0.5em}r{0.5em}){7-8}
\multicolumn{1}{c}{(Abbr.)} & \multicolumn{1}{c}{(\#Classes)} & & & & size & Width & Depth \\
\midrule
Indian Liver Patient Dataset (ILPD) & BC (2) & 579 & 10 & Yes & 64  & 512 & 1 \\
Heart Failure Clinical Records (HFC) & BC (2) & 299 & 12 & No & 64 & 512 & 5 \\
Cardiotocography (CTG) & NMC (10) & 2126 & 21 & No & 128 & 256 & 2 \\
Epileptic Seizure Recognition (ESR) & NMC (5) & 11500 & 178 & No & 256 & 512 & 4 \\
Estimation of Obesity Levels (EOL) & OMC (7) & 2111 & 16 & No & 128 & 128 & 2 \\
Maternal Health Risk (MHR) & OMC (3) & 1013 & 6 & No & 64 & 1024 & 4 \\
Parkinsons Telemonitoring (PT) & Reg (-) & 5875 & 19 & No & 128 & 1024 & 2 \\
Body Fat Prediction (BFP) & Reg (-) & 252 & 13 & No & 64 & 512 & 5 \\
\bottomrule
\end{tabular}%
}
\label{tab:dataset_summary}
\end{table*}

\noindent\textbf{\emph{Adaptive Binning as a Pretext Task.}} Numerical bin-index targets $y^{(n)}=B^{(n)}(x^{\mathrm{num}}_n)$ are refined online in a learning-driven, feature-wise coarse-to-fine curriculum. Each epoch minimizes $\Loss_{\mathrm{HORD}}$ (\emph{How}), whose per-feature numerical losses $\Loss_{\mathrm{num}}^{(n)}$ define the plateau metric $m_n$ used by FPT to trigger refinement upon saturation (\emph{When}); conditioned on a trigger, DIGS uses representations from uncorrupted inputs to split bins only when a candidate split improves both value-space variance reduction and representation-space coherence, yielding finer targets for subsequent epochs (\emph{Where}). Adaptive Binning (see Figure~\ref{figure:main}) therefore replaces a single global $T$ and fixed quantile boundaries with adaptive per-feature resolutions $\{T_n\}$, sharpening supervision without labels.

\section{Experiments and Results}
\noindent\textbf{Datasets.}
We curate a benchmark of publicly available medical tabular datasets spanning binary classification (BC), multiclass classification (MC; NMC for nominal, OMC for ordinal), and regression (Reg), with diverse clinical tasks, heterogeneous schemas, and varying categorical--numerical compositions (see Table~\ref{tab:dataset_summary}).

\noindent\textbf{Implementation Details.}
To ensure rigorous comparability, we adopt the 1000 epoch pretraining protocol of the fixed-binning baseline~\cite{lee2024binning}. As architectural complexity yields diminishing returns on tabular data~\cite{gorishniy2021revisiting,grinsztajn2022treebasedmodelsoutperformdeep,gorishniy2022embeddings}, we employ a standard MLP encoder $f_\theta$ with a symmetric decoder $f_d$. Dataset-specific depth $\in \{1,2,3,4,5\}$ and width $\in \{128,256,512,1024\}$ are selected via supervised validation (see Table~\ref{tab:dataset_summary}); all remaining configurations inherit from~\cite{lee2024binning}
\footnote{$\mathrm{lr}=10^{-4},10^{-2},10^{-3}$ for pretraining, linear probing, and fine-tuning, respectively.}.

\noindent\textbf{Evaluation.}
To isolate and quantify representation quality~\cite{lee2024binning,rubachev2022revisiting}, we employ two 100 epoch protocols: linear probing of frozen embeddings and fine-tuning. The fine-tuning phase pairs our encoder with MLPs and tabular architectures (ResNet, TabNet~\cite{arik2021tabnet}, FT-Transformer~\cite{gorishniy2021revisiting}, T2G-Former~\cite{yan2023t2g}), explicitly retaining their default configurations to minimize orthogonal influences from hyperparameter tuning. We report AUC for BC, Accuracy (Acc.) for NMC, QWK for OMC, and RMSE for Reg, all averaged over 10 seeds on a single NVIDIA RTX 4090.

\begin{table*}[t!]
\centering
\caption{
Linear evaluation with an MLP encoder across diverse datasets and tasks. We vary numerical binning (B: none (-), fixed binning (FIX), Ours), masking (M: none (-), constant replacement (C), random replacement (R)), and the pretext objective (O: ValueRecon (VR), MaskXent (MX), MaskXent+ValueRecon (MR), BinRecon (BR), Ours). We report Average Rank (Avg. Rank) aggregated from per-dataset rankings. Results are mean$_{\mathrm{std}}$; metrics are denoted as Task$_{\mathrm{metric}}$; \textBF{best} and \underline{second-best}.
}
{\fontsize{8pt}{9.6pt}\selectfont
\begin{tabular}{c c c c c c c c c c c c}
\toprule
\multicolumn{3}{c}{Datasets} &
  ILPD & HFC & CTG & ESR & EOL & MHR & PT & BFP & Avg. \\
\cmidrule(l{0.5em}r{0.5em}){4-5}\cmidrule(l{0.5em}r{0.5em}){6-7}\cmidrule(l{0.5em}r{0.5em}){8-9}\cmidrule(l{0.5em}r{0.5em}){10-11}
B & M & O &
  \multicolumn{2}{c}{$\text{BC}_{\text{AUC(\%) }\textcolor{red}{\uparrow}}$} &
  \multicolumn{2}{c}{$\text{NMC}_{\text{Acc.(\%) }\textcolor{red}{\uparrow}}$} &
  \multicolumn{2}{c}{$\text{OMC}_{\text{QWK(\%) }\textcolor{red}{\uparrow}}$} &
  \multicolumn{2}{c}{$\text{Reg}_{\text{RMSE }\textcolor{blue}{\downarrow}}$} &
  Rank \\
\midrule
- & - & VR & $75.89_{0.3}$ & $86.08_{2.4}$ & $85.66_{0.5}$ & $53.70_{0.2}$ & $86.38_{0.4}$ & $30.57_{1.2}$ & $15.98_{0.1}$ & $5.50_{0.0}$ & 10.88 \\
- & C & VR & $76.28_{0.1}$ & $85.09_{2.2}$ & $86.71_{0.3}$ & $57.00_{0.2}$ & $91.67_{0.4}$ & $41.97_{1.4}$ & $17.74_{0.2}$ & $5.53_{0.0}$ & \phantom{0}9.25 \\
- & C & MX & $76.52_{0.3}$ & $90.11_{0.5}$ & $83.52_{0.6}$ & $61.81_{0.2}$ & $87.05_{0.4}$ & $55.35_{0.8}$ & $19.57_{0.3}$ & $5.13_{0.1}$ & \phantom{0}8.19 \\
- & C & MR & $76.40_{0.2}$ & $84.85_{2.0}$ & $86.53_{0.3}$ & $58.72_{0.1}$ & $89.99_{0.3}$ & $38.91_{2.9}$ & $17.60_{0.2}$ & $5.64_{0.0}$ & \phantom{0}9.56 \\
- & R & VR & $76.39_{0.3}$ & $88.08_{2.1}$ & $86.46_{0.3}$ & $55.54_{0.2}$ & $91.91_{0.4}$ & $40.67_{2.1}$ & $17.40_{0.1}$ & $5.48_{0.0}$ & \phantom{0}8.38 \\
- & R & MX & $76.53_{0.3}$ & $89.19_{0.4}$ & $84.84_{0.3}$ & $62.67_{0.1}$ & $86.27_{0.3}$ & $64.06_{1.0}$ & $15.29_{0.3}$ & $5.45_{0.1}$ & \phantom{0}7.31 \\
- & R & MR & $76.48_{0.2}$ & $85.82_{2.1}$ & $86.53_{0.2}$ & $57.39_{0.2}$ & $90.66_{0.6}$ & $43.47_{1.7}$ & $15.27_{0.1}$ & $5.37_{0.1}$ & \phantom{0}7.31 \\
\midrule
FIX & - & BR & $75.54_{0.2}$ & $86.76_{0.3}$ & $84.86_{0.5}$ & $62.00_{0.1}$ & $86.83_{0.4}$ & $60.45_{0.5}$ & $16.67_{0.1}$ & $5.32_{0.1}$ & \phantom{0}8.88 \\
FIX & C & BR & $76.25_{0.3}$ & $90.11_{0.4}$ & $86.76_{0.2}$ & $62.94_{0.2}$ & $90.52_{0.4}$ & $61.03_{0.5}$ & $17.66_{0.2}$ & $5.30_{0.0}$ & \phantom{0}6.31 \\
FIX & R & BR & $76.10_{0.3}$ & $86.34_{0.5}$ & $85.89_{0.3}$ & $63.83_{0.2}$ & $88.13_{0.8}$ & $62.41_{0.4}$ & $15.71_{0.4}$ & $5.33_{0.0}$ & \phantom{0}7.38 \\
\midrule
\rowcolor[gray]{.93}\textBF{Ours} & - & \textBF{Ours} & $76.53_{0.1}$ & $93.25_{0.6}$ & $84.91_{0.2}$ & $64.10_{0.2}$ & $\textBF{93.78}_{0.4}$ & $64.13_{0.9}$ & $\underline{14.27}_{0.1}$ & $4.65_{0.0}$ & \phantom{0}3.56 \\
\rowcolor[gray]{.93}\textBF{Ours} & C & \textBF{Ours} & $\textBF{77.80}_{0.2}$ & $\underline{95.00}_{0.3}$ & $\textBF{87.70}_{0.5}$ & $\textBF{66.86}_{0.1}$ & $89.71_{0.5}$ & $\underline{66.91}_{0.8}$ & $14.98_{0.1}$ & $\underline{4.57}_{0.0}$ & \phantom{0}\underline{2.50} \\
\rowcolor[gray]{.93}\textBF{Ours} & R & \textBF{Ours} & $\underline{77.25}_{0.2}$ & $\textBF{96.88}_{0.3}$ & $\underline{87.61}_{0.4}$ & $\underline{65.60}_{0.1}$ & $\underline{93.17}_{0.4}$ & $\textBF{70.51}_{1.2}$ & $\textBF{11.32}_{0.1}$ & $\textBF{4.56}_{0.0}$ & \textBF{\phantom{0}1.50} \\
\bottomrule
\end{tabular}%
}
\label{tab:linear_main}
\end{table*}

\noindent\textBF{Linear Evaluation.} Table~\ref{tab:linear_main} supports a clear takeaway on medical tabular tasks: adaptive discretization achieves the best average rank across masking options and pretext objectives. Under masking, we tune $p_m\in\{0.1,0.2,0.3\}$ and choose the initial resolution (fixed $T$ for BinRecon, $T_{\mathrm{init}}$ for ours) from $\{2,10\}$. Notably, the margin over fixed-binning BinRecon~\cite{lee2024binning} persists even at its best masked setting, and our no-mask variant still surpasses masked BinRecon. This pattern indicates that improvements are driven primarily by training-adaptive, feature-wise refinement rather than input corruption, with masking acting as a complementary regularizer. Overall, the \emph{When--Where--How} coupling turns discretization into a pretext that adaptively sharpens supervision and yields stronger representations for clinical tables.

To assess the contribution of each proposed component, Table~\ref{tab:ablation} reveals a compositional pattern. Ablating any single component degrades linear probing, indicating complementary effects that are not attributable to one module alone. The HF setting is especially instructive. Because FPT never triggers, supervision remains effectively fixed-binned, yet removing HORD still induces a marked drop, demonstrating the value of type-aware ordinal supervision even without refinement. Overall, the ablations indicate that performance is driven by the integrated refinement pipeline.

\begin{table*}[t!]
\centering
\caption{
Linear-evaluation ablations of our method across datasets.
Each variant removes one component from the full model (w/o FWA/HORD: direct removal; w/o FPT: DIGS with fixed-epoch refinement; w/o DIGS: FPT with variance-only splitting); all other settings follow Table~\ref{tab:linear_main}. Mean$_{\mathrm{std}}$; metrics as Task$_{\mathrm{metric}}$; best in \textBF{bold}.
}
{\fontsize{8pt}{9.6pt}\selectfont
\begin{tabular}{c cc cc cc cc}
\toprule
\multirow{-1}{*}{Datasets} &
ILPD & HFC & CTG & ESR & EOL & MHR & PT & BFP \\
\cmidrule(l{0.5em}r{0.5em}){2-3}\cmidrule(l{0.5em}r{0.5em}){4-5}\cmidrule(l{0.5em}r{0.5em}){6-7}\cmidrule(l{0.5em}r{0.5em}){8-9}
Method &
\multicolumn{2}{c}{BC$_{\text{AUC(\%) }\textcolor{red}{\uparrow}}$} &
\multicolumn{2}{c}{NMC$_{\text{Acc.(\%) }\textcolor{red}{\uparrow}}$} &
\multicolumn{2}{c}{OMC$_{\text{QWK(\%) }\textcolor{red}{\uparrow}}$} &
\multicolumn{2}{c}{Reg$_{\text{RMSE }\textcolor{blue}{\downarrow}}$}
\\
\midrule
\rowcolor[gray]{.93}\multicolumn{1}{c}{\textBF{Ours}} &
\multicolumn{1}{l}{$\textBF{77.80}_{0.2}$} &
\multicolumn{1}{l}{$\textBF{96.88}_{0.3}$} &
\multicolumn{1}{l}{$\textBF{87.70}_{0.5}$} &
\multicolumn{1}{l}{$\textBF{66.86}_{0.1}$} &
\multicolumn{1}{l}{$\textBF{93.78}_{0.4}$} &
\multicolumn{1}{l}{$\textBF{70.51}_{1.2}$} &
\multicolumn{1}{l}{$11.32_{0.1}$} &
\multicolumn{1}{l}{$\textBF{4.56}_{0.0}$} \\
\multicolumn{1}{l}{w/o FWA} &
$76.73_{0.2}$ &
$\textBF{96.88}_{0.3}$ &
$85.02_{0.4}$ &
$63.97_{0.1}$ &
$89.15_{0.3}$ &
$67.07_{0.6}$ &
$\textBF{11.23}_{0.1}$ &
$4.97_{0.0}$ \\
\multicolumn{1}{l}{w/o FPT} &
$76.48_{0.3}$ &
$\textBF{96.88}_{0.3}$ &
$86.50_{0.3}$ &
$64.31_{0.2}$ &
$84.08_{0.5}$ &
$66.23_{1.3}$ &
$14.20_{0.2}$ &
$4.91_{0.1}$ \\
\multicolumn{1}{l}{w/o DIGS} &
$74.07_{0.3}$ &
$\textBF{96.88}_{0.3}$ &
$86.29_{0.4}$ &
$65.97_{0.1}$ &
$85.59_{0.4}$ &
$67.29_{0.8}$ &
$13.55_{0.1}$ &
$5.44_{0.1}$ \\
\multicolumn{1}{l}{w/o HORD} &
$76.51_{0.2}$ &
$88.41_{0.3}$ &
$86.71_{0.5}$ &
$62.80_{0.1}$ &
$91.39_{0.6}$ &
$63.31_{1.1}$ &
$15.84_{0.2}$ &
$5.12_{0.0}$ \\
\bottomrule
\end{tabular}
}
\label{tab:ablation}
\end{table*}

\begin{figure}[t!]
\begin{center}
\includegraphics[width=\columnwidth]{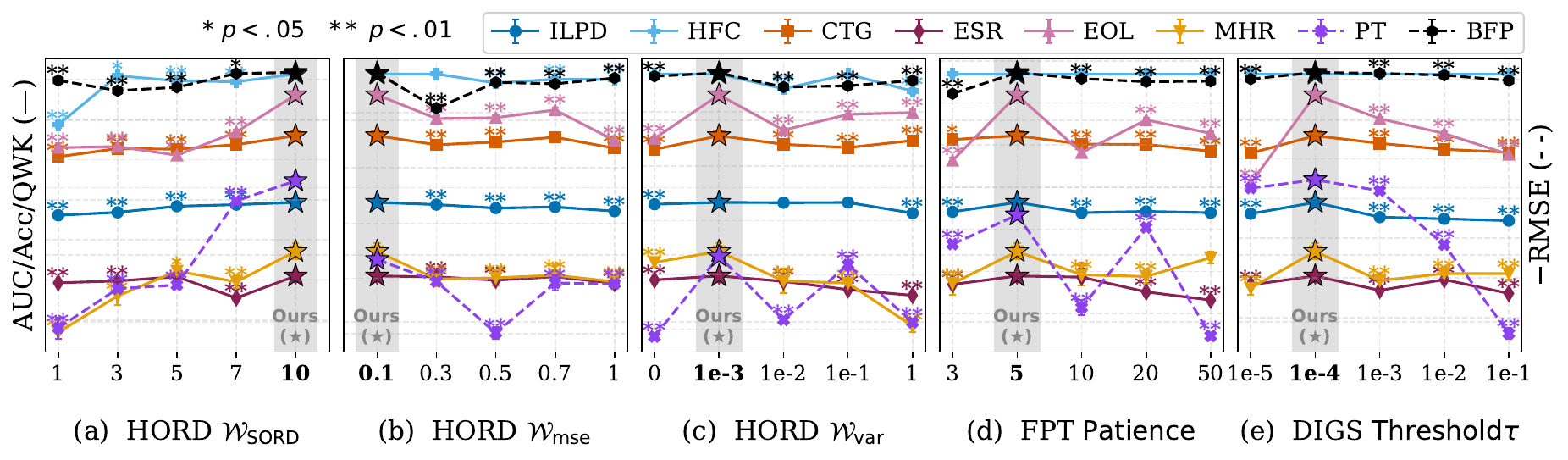}
\caption{
Linear-probing hyperparameter sweeps under the same setup as Table~\ref{tab:linear_main}.
We sweep (a) $w_{\mathrm{SORD}}\in\{1,3,5,7,\textBF{10}\}$, (b) $w_{\mathrm{MSE}}\in\{\textBF{0.1},0.3,0.5,0.7,1\}$, (c) $w_{\mathrm{Var}}\in\{0,\textBF{10}^{\textbf{-3}},10^{-2},10^{-1},1\}$, (d) FPT patience $\in\{3,\textBF{5},10,20,50\}$, and (e) DIGS threshold $\tau\in\{10^{-5},\textBF{10}^{\textbf{-4}},10^{-3},10^{-2},10^{-1}\}$.
The gray shaded line denotes the default configuration; statistical significance markers are shown above each point. 
}
\label{figure:loss_sensitivity}
\end{center}
\end{figure}

Finally, Figure~\ref{figure:loss_sensitivity} demonstrates the method's robustness to hyperparameter choice. A single default configuration provides a reliable starting point across tasks and datasets, reducing the need for per-dataset tuning, where extensive tuning can amplify the risk in clinical deployment~\cite{varoquaux2022machine}. Across broad sweeps of loss weights and refinement controls, deviations from the default consistently reduce performance, reinforcing it as a robust choice.

\begin{table*}[t!]
\centering
\caption{
Fine-tuning with tabular-specific encoders: supervised from scratch vs.\ SSL-pretrained initialization.
MaskXent+ValueRecon (MR) and fixed-binning BinRecon (BR) serve as SSL baselines based on their strong linear-probe results (Table~\ref{tab:linear_main}).
Results are mean$_{\mathrm{std}}$ across runs; metrics are denoted as Task$_{\mathrm{metric}}$; \textbf{best} and \underline{second-best}.
}
{\fontsize{8pt}{9.5pt}\selectfont
\begin{tabular}{c cc c c c c}
\toprule
\multirow{2}{*}{Encoder} &
\multicolumn{2}{c}{Datasets} &
HFC & ESR & EOL & BFP \\
\cmidrule(l{0.5em}r{0.5em}){4-4}\cmidrule(l{0.5em}r{0.5em}){5-5}\cmidrule(l{0.5em}r{0.5em}){6-6}\cmidrule(l{0.5em}r{0.5em}){7-7}
&
\multicolumn{2}{c}{Method} &
BC$_{\text{AUC(\%) }\textcolor{red}{\uparrow}}$ &
NMC$_{\text{Acc.(\%) }\textcolor{red}{\uparrow}}$ &
OMC$_{\text{QWK(\%) }\textcolor{red}{\uparrow}}$ &
Reg$_{\text{RMSE }\textcolor{blue}{\downarrow}}$ \\
\midrule
\multirow{3}{*}{MLP}
& \multicolumn{2}{c}{Supervised} & $\underline{90.74}_{1.6}$ & $76.24_{0.5}$ & $94.06_{0.6}$ & $5.50_{0.3}$ \\
& SSL & MR & $89.90_{1.2}$ & $76.66_{0.6}$ & $\underline{94.64}_{0.6}$ & $\underline{5.40}_{0.2}$ \\
& SSL & BR & $90.54_{0.7}$ & $\textBF{77.02}_{0.6}$ & $\textBF{94.74}_{0.6}$ & $5.65_{0.3}$ \\
& \cellcolor[gray]{.93}SSL & \cellcolor[gray]{.93}\textBF{Ours} & \cellcolor[gray]{.93}$\textBF{90.87}_{0.9}$ & \cellcolor[gray]{.93}$\underline{76.87}_{0.6}$ & \cellcolor[gray]{.93}$93.89_{0.9}$ & \cellcolor[gray]{.93}$\textBF{5.09}_{0.2}$ \\
\midrule
\multirow{3}{*}{ResNet}
& \multicolumn{2}{c}{Supervised} & $\textBF{91.12}_{1.6}$ & $\underline{76.33}_{0.7}$ & $88.22_{1.3}$ & $5.43_{0.4}$ \\
& SSL & MR & $\underline{90.97}_{0.6}$ & $76.27_{0.3}$ & $88.81_{0.8}$ & $5.18_{0.2}$ \\
& SSL & BR & $90.96_{0.9}$ & $75.95_{0.5}$ & $\underline{88.89}_{1.3}$ & $\underline{4.82}_{0.2}$ \\
& \cellcolor[gray]{.93}SSL & \cellcolor[gray]{.93}\textBF{Ours} & \cellcolor[gray]{.93}$90.52_{1.6}$ & \cellcolor[gray]{.93}$\textBF{76.70}_{0.3}$ & \cellcolor[gray]{.93}$\textBF{89.17}_{1.0}$ & \cellcolor[gray]{.93}$\textBF{4.77}_{0.1}$ \\
\midrule
\multirow{3}{*}{TabNet~\cite{arik2021tabnet}}
& \multicolumn{2}{c}{Supervised} & $85.07_{7.3}$ & $47.89_{1.8}$ & $64.51_{6.8}$ & $10.22_{8.0}$ \\
& SSL & MR & $87.22_{4.3}$ & $48.82_{2.4}$ & $62.69_{5.1}$ & $9.40_{1.7}$ \\
& SSL & BR & $\underline{87.39}_{2.7}$ & $\underline{50.88}_{1.8}$ & $\underline{74.26}_{4.8}$ & $\phantom{0}\underline{8.42}_{1.6}$ \\
& \cellcolor[gray]{.93}SSL & \cellcolor[gray]{.93}\textBF{Ours} & \cellcolor[gray]{.93}$\textBF{89.31}_{2.5}$ & \cellcolor[gray]{.93}$\textBF{52.32}_{1.6}$ & \cellcolor[gray]{.93}$\textBF{75.41}_{4.3}$ & \cellcolor[gray]{.93}$\phantom{0}\textBF{8.23}_{1.4}$ \\
\midrule
\multirow{3}{*}{FT-Transformer~\cite{gorishniy2021revisiting}}
& \multicolumn{2}{c}{Supervised} & $89.43_{3.5}$ & $67.16_{1.0}$ & $\textBF{95.78}_{1.0}$ & $5.52_{0.1}$ \\
& SSL & MR & $92.21_{2.4}$ & $70.39_{0.6}$ & $\underline{94.37}_{1.1}$ & $\underline{5.27}_{0.2}$ \\
& SSL & BR & $\underline{92.47}_{2.1}$ & $\underline{74.14}_{0.3}$ & $93.62_{1.2}$ & $5.70_{0.3}$ \\
& \cellcolor[gray]{.93}SSL & \cellcolor[gray]{.93}\textBF{Ours} & \cellcolor[gray]{.93}$\textBF{93.43}_{1.5}$ & \cellcolor[gray]{.93}$\textBF{75.97}_{0.7}$ & \cellcolor[gray]{.93}$93.53_{1.4}$ & \cellcolor[gray]{.93}$\textBF{5.05}_{0.3}$ \\
\midrule
\multirow{3}{*}{T2G-Former~\cite{yan2023t2g}}
& \multicolumn{2}{c}{Supervised} & $91.77_{3.7}$ & $67.93_{0.0}$ & $96.67_{1.4}$ & $\textBF{5.01}_{0.1}$ \\
& SSL & MR & $92.90_{1.5}$ & $\underline{72.62}_{0.9}$ & $96.75_{0.9}$ & $5.32_{0.2}$ \\
& SSL & BR & $\underline{93.17}_{2.0}$ & $71.90_{0.0}$ & $\textBF{97.09}_{1.3}$ & $5.27_{0.3}$ \\
& \cellcolor[gray]{.93}SSL & \cellcolor[gray]{.93}\textBF{Ours} & \cellcolor[gray]{.93}$\textBF{94.22}_{1.7}$ & \cellcolor[gray]{.93}$\textBF{73.12}_{0.0}$ & \cellcolor[gray]{.93}$\textBF{97.09}_{0.9}$ & \cellcolor[gray]{.93}$\underline{5.02}_{0.3}$ \\
\bottomrule
\end{tabular}%
}
\label{tab:finetune}
\end{table*}

\noindent\textBF{Fine-tuning Evaluation.} Table~\ref{tab:finetune} shows that the benefits of our pretraining persist under end-to-end fine-tuning, rather than being confined to linear probing. Among SSL objectives, adaptive discretization provides the most reliable initialization and typically improves over MR and fixed-binning BR after fine-tuning. While purely supervised training can be optimal for particular model--task pairs, the overall trend indicates a robustness advantage: our pretraining reaches competitive or superior optima with reduced sensitivity to the downstream model choice, which is valuable given diverse encoder choices in medical tabular modeling. Overall, these results suggest that learning-driven discretization acts as a transferable inductive bias that persists under downstream optimization, rather than a probe-specific artifact.

\section{Conclusion}
We introduce Adaptive Binning, a tabular self-supervised pretext for medical data that elevates discretization from a fixed design choice to a learning-coupled, feature-wise coarse-to-fine curriculum. By designing plateau-triggered refinement, representation-aware split selection, and heterogeneity-aware ordinal supervision, our approach yields stronger representations across tasks and datasets. We further establish a benchmark of medical tabular datasets with unified evaluation protocols, enabling reproducible comparisons for self-supervised learning on clinical tables.  Our study is limited to in-dataset transfer and a small set of downstream protocols; future work will extend evaluation to broader clinical endpoints and cross-dataset pretraining with adaptation to new targets.

\begin{credits}

\subsubsection{\discintname}
The authors have no competing interests to declare.
\end{credits}

\bibliographystyle{splncs04}
\bibliography{main}

\end{document}